%% file: root.tex
\documentclass[letterpaper, 10 pt, conference]{ieeeconf}  

\IEEEoverridecommandlockouts                              

\usepackage{xcolor}
\usepackage{graphicx}
\usepackage[caption=false,font=footnotesize]{subfig}
\usepackage{hyperref}
\usepackage{amsmath}
\usepackage{amssymb}
\usepackage{amsfonts}
\usepackage{scrextend}
\usepackage{mathtools}
\usepackage{booktabs}
\usepackage{tabulary}
\usepackage{multirow}
\usepackage{tikz}

\usetikzlibrary{calc}
\usetikzlibrary{shapes.geometric}
\usetikzlibrary{fit}
\usetikzlibrary{positioning}
\usetikzlibrary{decorations.pathmorphing, decorations.pathreplacing,angles,quotes}
\usetikzlibrary{arrows}

\input{format/math_operators}

\title{\LARGE \bf
See What the Robot Can’t See:\\Learning Cooperative Perception for Visual Navigation 
}

\author{Jan Blumenkamp\textsuperscript{*}\hspace{8pt}%
Qingbiao Li\textsuperscript{*}\hspace{8pt}%
Binyu Wang\hspace{8pt}%
Zhe Liu\textsuperscript{*}\hspace{8pt}%
Amanda Prorok\textsuperscript{*}
\thanks{$^{*}$J. Blumenkamp, Q. Li, Z. Liu and A. Prorok are with the Department of Computer Science and Technology, University of Cambridge, United Kingdom
        {\tt\small \{jb2270, ql295, zl457, asp45\}@cam.ac.uk}}%
}

\begin{document}

\maketitle
\thispagestyle{empty}
\pagestyle{empty}

\input{main}

\bibliographystyle{IEEEtran}
\bibliography{references}

\end{document}

%% file: format/math_operators.tex
\newcommand{\expect}[2]{\mathbb{E}_{#1}\left[#2\right]}

\newcommand{\bbm}{\begin{bmatrix}}
\newcommand{\ebm}{\end{bmatrix}}

%% file: main.tex
\begin{abstract}
We consider the problem of navigating a mobile robot towards a target in an unknown environment that is endowed with visual sensors, where neither the robot nor the sensors have access to global positioning information and only use first-person-view images.
In order to overcome the need for positioning, we train the sensors to encode and communicate relevant viewpoint information to the mobile robot, whose objective it is to use this information to navigate to the target along the shortest path.
We overcome the challenge of enabling \textit{all} the sensors (even those that cannot directly see the target) to predict the direction along the shortest path to the target by implementing a neighborhood-based feature aggregation module using a Graph Neural Network (GNN) architecture.
In our experiments, we first demonstrate generalizability to previously unseen environments with various sensor layouts. Our results show that by using \textit{communication} between the sensors and the robot, we achieve up to $2.0\times$ improvement in SPL (Success weighted by Path Length) when compared to a \textit{communication-free} baseline. This is done without requiring a global map, positioning data, nor pre-calibration of the sensor network. 
Second, we perform a zero-shot transfer of our model from simulation to the real world. 
Laboratory experiments demonstrate the feasibility of our approach in various cluttered environments. Finally, we showcase examples of successful navigation to the target while both the sensor network layout as well as obstacles are dynamically reconfigured as the robot navigates. We provide a video demo\footnote{\scriptsize{\url{https://www.youtube.com/watch?v=kcmr6RUgucw}}}, the dataset, trained models, and source code
\footnote{\scriptsize{\url{https://github.com/proroklab/sensor-guided-visual-nav}}}.
\end{abstract}


\section{Introduction}
	
Efficiently finding and navigating to a target in complex unknown environments is a fundamental robotics problem, with applications to search and rescue~\cite{Mirowski2017}, surveillance~\cite{Kolling2009}, and monitoring~\cite{Corke2005}. 
We focus on navigating in an unknown environment to find a target under the guidance of a \textit{visual sensor network}. Prior work has provided effective solutions employing low-cost wireless sensors to guide robotic navigation~\cite{Corke2005,Kashino2020}. 
These studies demonstrate that at a small additional cost---i.e., the deployment of cheap sensors with local communication capabilities---the requirements on the robot's capabilities can be significantly reduced while simultaneously improving its navigation efficiency.

However, the implementation of traditional sensor network guided navigation is cumbersome, as it commonly requires that \textit{(i)} robots and sensors have access to external positioning systems (e.g., GPS), and that \textit{(ii)} an explicit map is built (that the robot can then use to plan a path). 
This approach has as a main drawback that sensors and robots are required to accurately estimate their positions within a global frame of reference at all times (e.g., through an available global positioning system such as GPS). Also, it requires the design of a multi-hop/multi-round communication strategy that allows the sensors to jointly build and maintain an up-to-date map of the environment.

\begin{figure}[t]
    \centering
    \includegraphics[width=\linewidth]{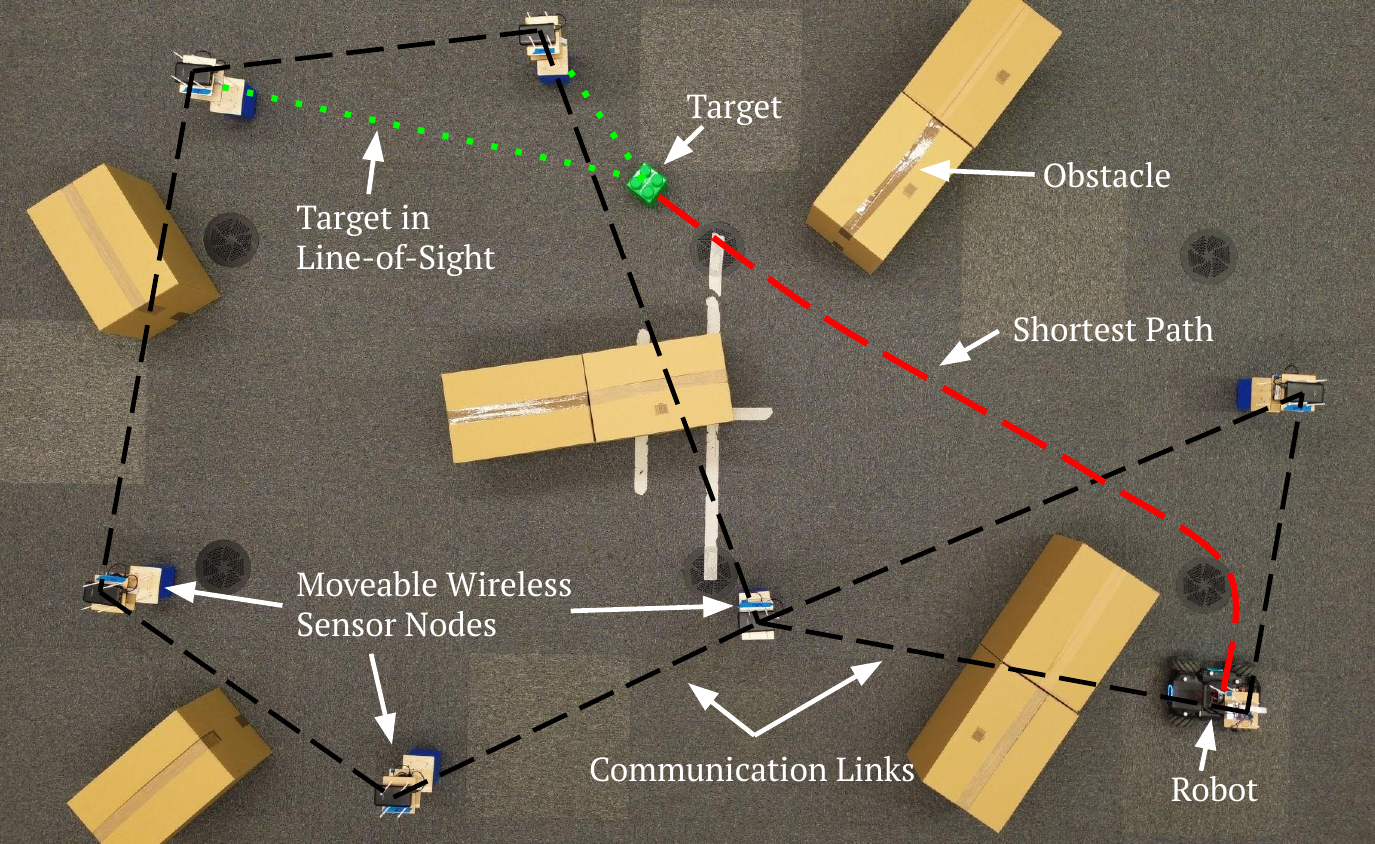}
    \caption{Our setup consists of a workspace populated with a target, obstacles, a mobile robot, and movable wireless sensor nodes equipped with cameras. We train a GNN (Graph Neural Network) to guide the robots toward the target along the shortest path. Our setup is positioning-free, relies on visual information and communication only, and therefore allows for dynamic online reconfiguration of all components.}
\label{fig:hero}
\end{figure}

In this paper, we consider the problem of using a \textit{positioning-free} visual sensor network that \textit{learns} to guide the robot to its target.
Successful navigation requires the robot to aggregate and process information received from nearby sensors, helping it decide in which direction to move. Towards this end, we propose a data-driven approach wherein the nodes in the network learn a policy that incorporates both visual processing as well as a communication module.
We propose a Graph Neural Network (GNN)-based architecture that learns to encode and communicate relevant viewpoint information to the mobile robot. During navigation, the robot is guided by a model that we train through imitation learning to predict the effective cost-to-go (to the target), thereby approximating optimal motion primitives. The setup is depicted in \autoref{fig:hero}.
Our experiments demonstrate successful sim-to-real transfer to a laboratory setup. We also showcase the ability of our network to adapt to dynamic changes in its configuration by showing examples
of successful navigation to the target while sensors nodes are being moved.

\begin{figure*}[bt]
    \centering
    \subfloat[Map]{\includegraphics[height=2.9cm]{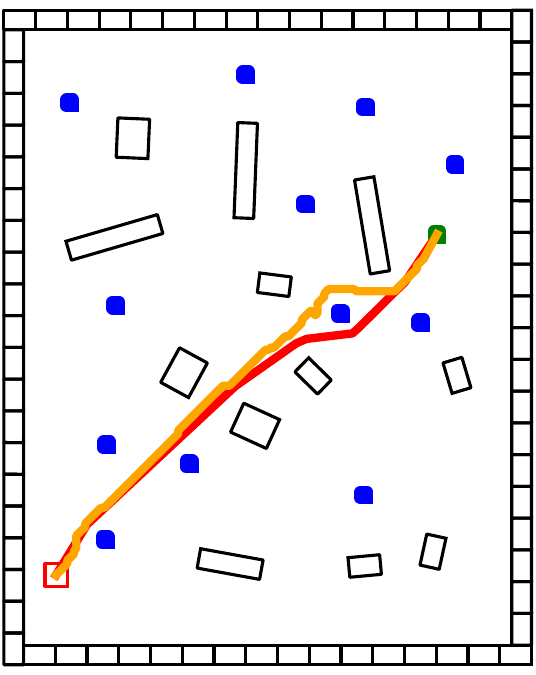}}\label{fig:large_env}\hfill
    \subfloat[Webots]{\includegraphics[height=2.9cm]{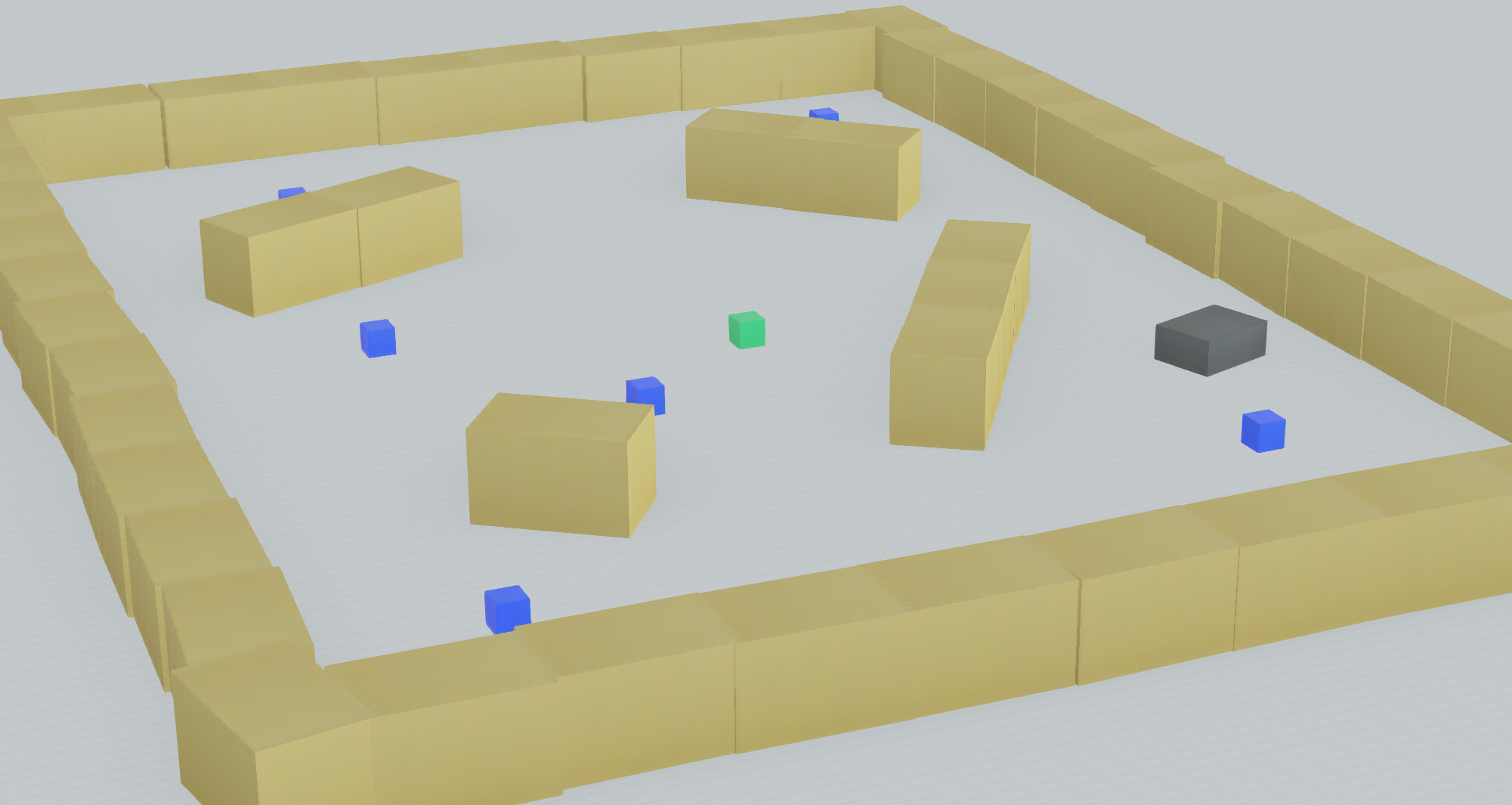}}\label{fig:env_sim}\hfill
    \subfloat[Real world]{\includegraphics[height=2.9cm]{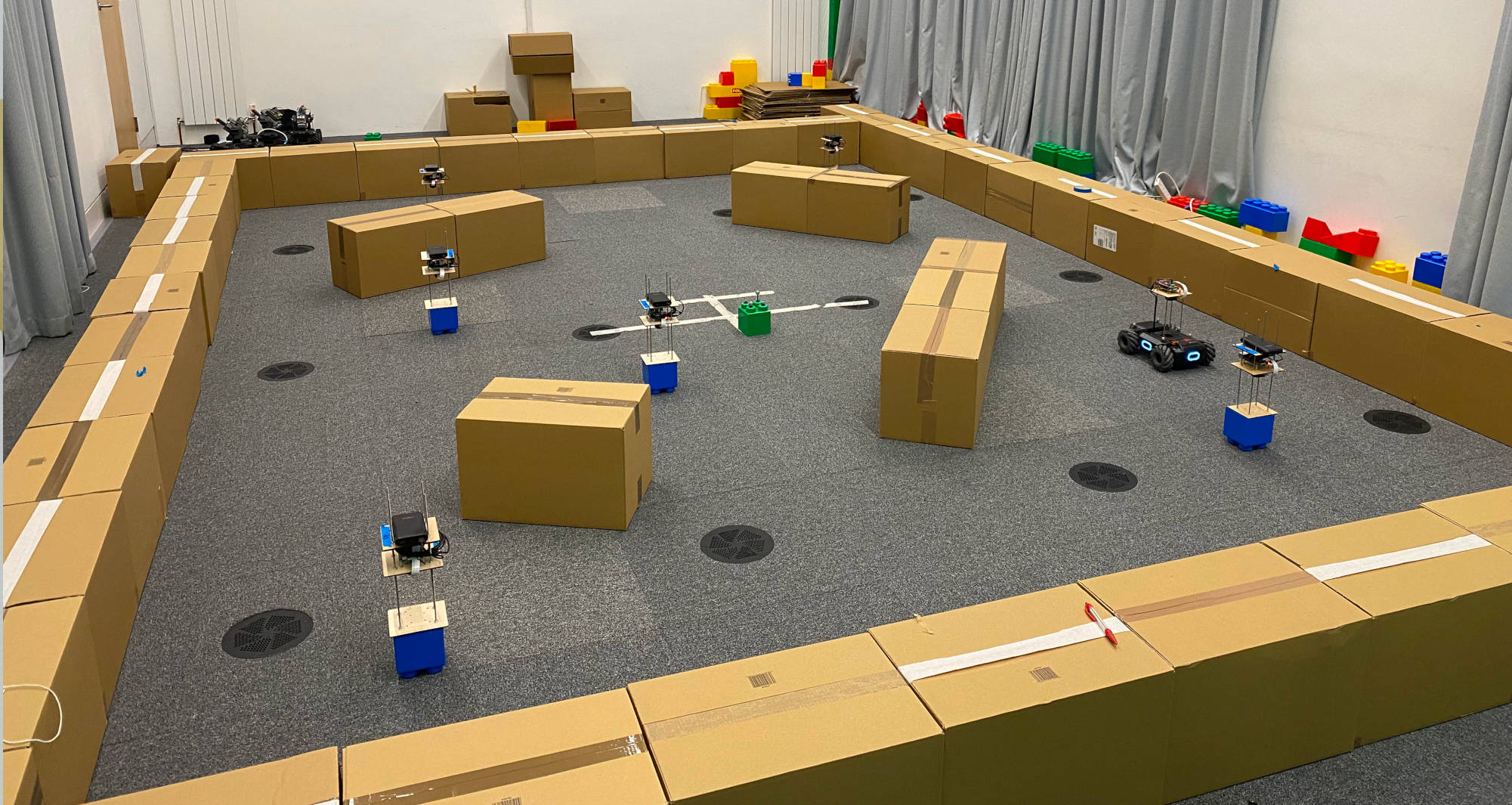}}\label{fig:env_real}\hfill
    \subfloat[Sensor]{\includegraphics[height=2.9cm]{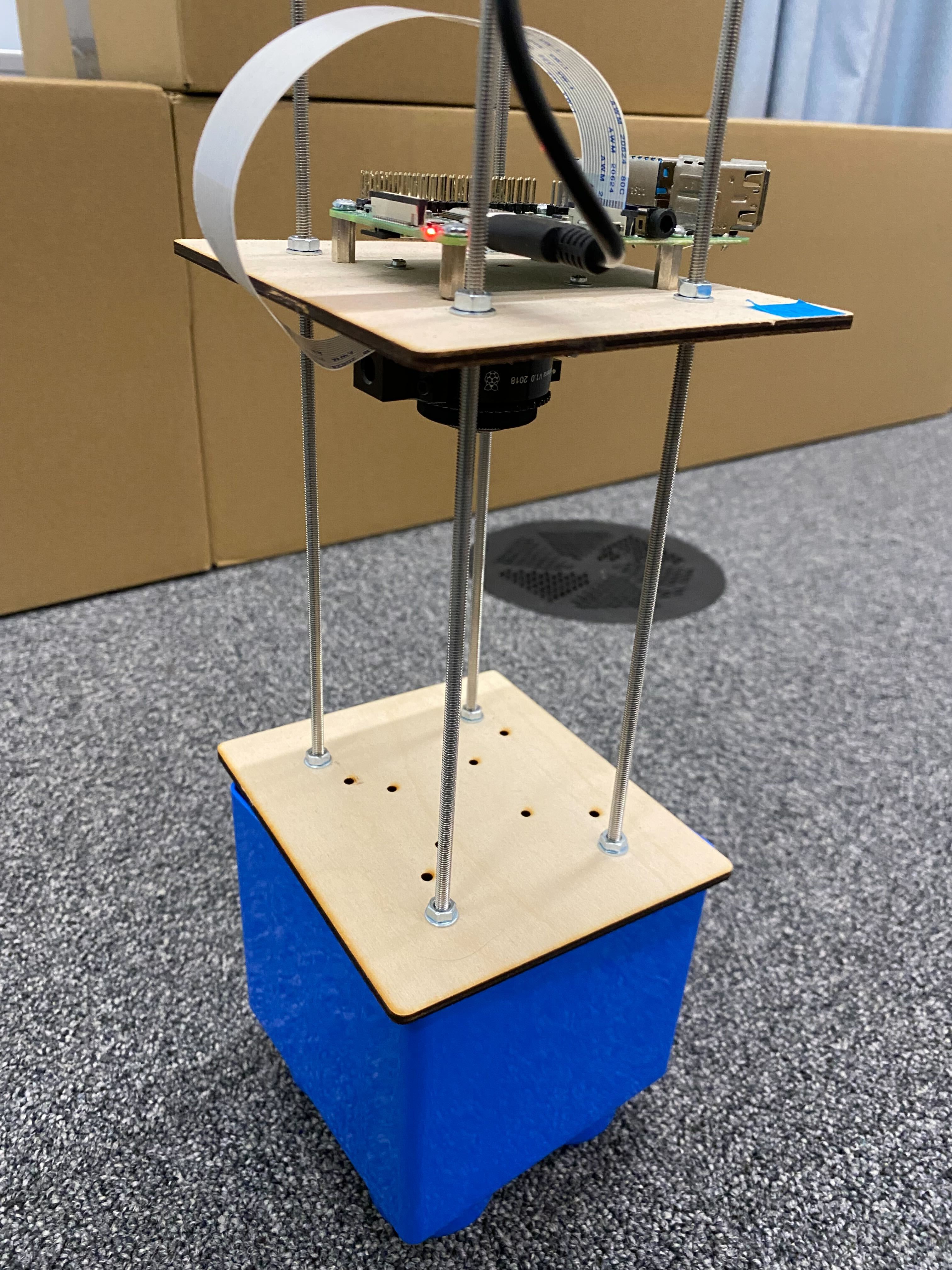}}\label{fig:sensor}\hfill
    \subfloat[Robot]{\includegraphics[height=2.9cm]{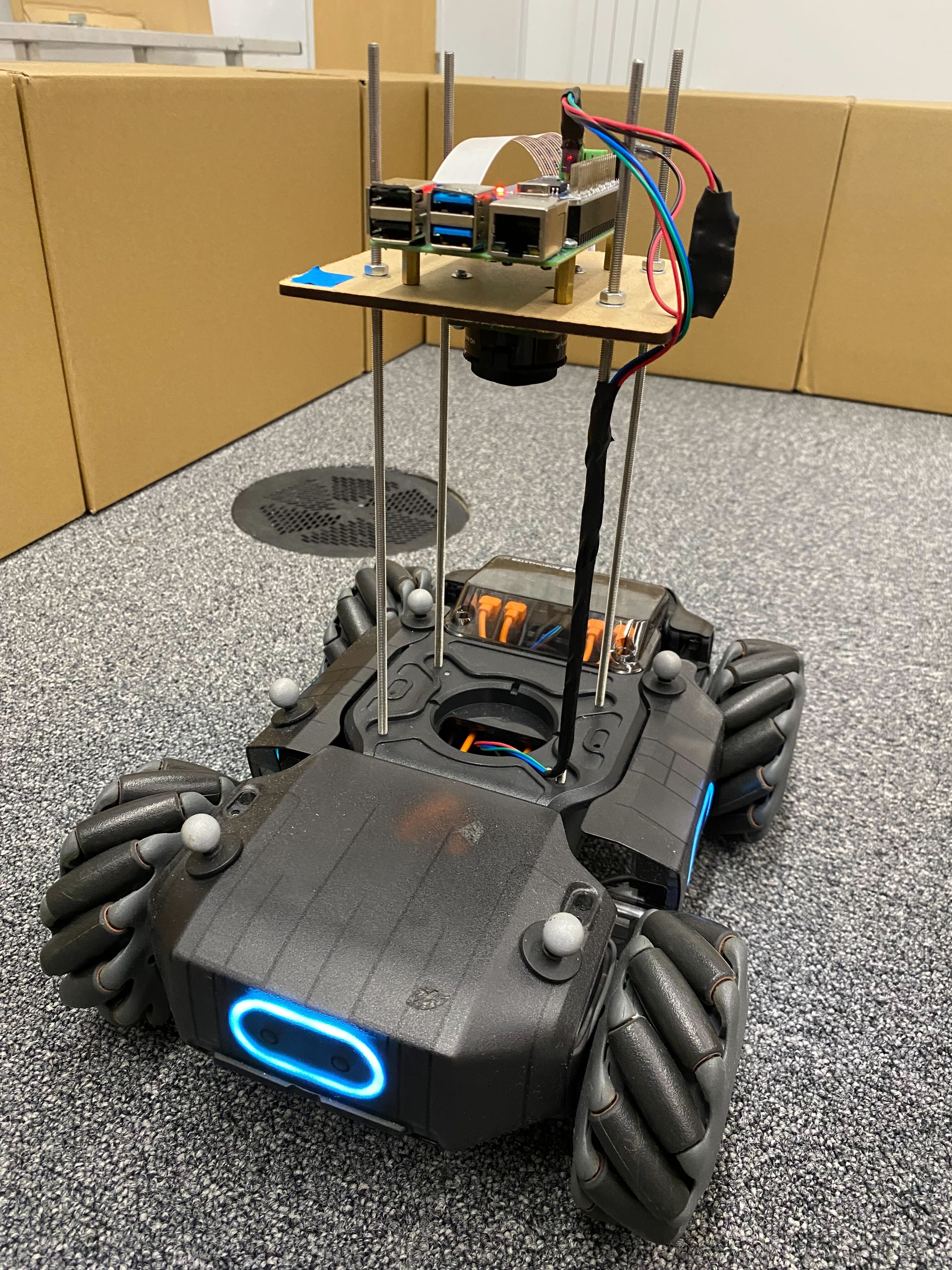}}\label{fig:robot}
    \caption{Our setup consists of (a) an environment populated with obstacles, visual sensors (blue), a mobile robot equipped with a sensor, and a green target, where the mobile robot has to find the shortest path (red, taken path orange) to an occluded target by incorporating communicated sensor information to enhance its navigation performance. Our framework leverages a setup consisting of a simulation environment (b) that corresponds to the real-world setup (c). The workspace is endowed with custom-built sensors with fish-eye cameras (d, e) that are capable of communicating with each other. One sensor is attached to a mobile robot (e) that acts on the output of that sensor.}
\label{fig:system}
\end{figure*}

\subsection{Related Work}

{\it Sensor Network-Guided Navigation.}
Early approaches assume that either the robot~\cite{Corke2005,Shenoy2005} or the sensors~\cite{Batalin2004, li2003distributed} are in the same absolute reference frame to create an explicit environment map to be used by conventional planners.
Decentralized first-principles based vision-based communication-free approaches~\cite{shah2019swarm} or Gaussian Belief Propagation \cite{Ortiz2021visualGBP} and Factor Graphs \cite{gtsam} are used for decentralized robot mapping, using probabilistic models to estimate information such as the position of nodes in a graph using local information. These methods require human prior knowledge to design the model and to extract relevant local features. In contrast, our method is completely data-driven and trained end-to-end using first-person view visual information and thus promises to scale to complex real-world scenes.
Deep Learning (DL)-based methods~\cite{Bhatti2018} are becoming more attractive, but still require anchor-nodes with access to global positioning information.

{\it GNNs for sensor networks and mobile robot systems.} As an effective method to aggregate and learn from relational, non-Euclidean data, GNNs have achieved promising results in numerous domains~\cite{Akter2019} such as multi-robot path planning~\cite{Li2020} or coverage~\cite{Jan2020}.
VGAI \cite{hu2021vgai} is the first to use first-person view visual information to imitate a flocking policy in a swarm of drones using GNNs. In contrast to visual navigation, information from the close neighborhood is sufficient for a reasonable flocking policy, whereas for target-navigation, information about the direction to the target has to propagate through the network and reach the navigating robot. Furthermore, our paper is the first to demonstrate a GNN-based and vision-based real-world deployment.

{\it Imitation Learning for Navigation.}
\textsc{Dagger} is able to learn a stationary deterministic policy guaranteed to perform well under its induced distribution of states by using reduction-based approaches \cite{ross_dagger}. \textsc{AggreVate} improves further on this by iteratively estimating a cost-to-go function \cite{ross_aggrevate}. Traditional approaches for navigation in unknown environments usually utilize the greedy search with Euclidean heuristic or Manhattan heuristic, which are inefficient in detecting and escaping local minima \cite{pmlr-v78-bhardwaj17a}. Recent learning-based approaches imitate oracles to offer better sample efficiency \cite{7487175}.
In this paper, we employ a visual sensor network to implicitly learn the cost-to-go by predicting the direction corresponding to the shortest path to the target computed by an expert planner.

{\it Sim-to-Real Transfer and Real-World GNN Implementation.}
The gap between the simulator and the real world, where dynamics and vision differ, makes it difficult to transfer learned policy. 
Image-to-image translation has been used for tactile sensors~\cite{isola2017image} or for robot control~\cite{zhang2019vr} for sim-to-real transfer, by transferring image from the real-world to the simulation domain, often using architectures from the CycleGAN~\cite{zhu2017unpaired} family. In this paper, we leverage GNNs for robot navigation tasks. 
Recent work \cite{Blumenkamp2022} builds the first real-world deployment of GNN-based policies on a decentralized multi-robot platform, but relies on global positioning and a relatively simple state space.

\subsection{Contributions}
Our contributions include the following:
\begin{itemize}
\item We present a framework that demonstrates, for the first time, how low-cost sensor networks can help robots navigate to targets in unknown environments, \textit{where neither the robot nor the sensors possess any absolute or relative positioning information}.
    
    \item We provide an end-to-end visual navigation policy that leverages GNNs to \textit{learn what needs to be communicated} (among sensors and robot) and how to aggregate the visual scene for effective navigation.
    
    \item Experimental results demonstrate generalizability to unseen environments with various sensor layouts. In particular, by introducing a \textit{real-to-sim image translator}, our policy (which is trained entirely in simulation) can be transferred to the real world without additional tuning (i.e., in a zero-shot manner).
\end{itemize}

\section{Problem Formulation}
\label{sec:problem_formulation}

Our setup consists of a cluttered environment, a mobile robot, and a set of visual sensors. The continuous environment $\mathcal{W}$ contains a set of randomly placed obstacles $\mathcal{C}\subset\mathcal{W}$ and $N$ randomly placed visual sensors $\mathcal{S}=\{S^1, \dots, S^N\}$. As shown in \autoref{fig:system} (c), at every time step $t$, each sensor $S^i$ is capable of taking an omnidirectional RGB image $o_t^i$ of its surrounding environment (equirectangular projection of a fisheye lens), but has \textit{no positioning information}. Each sensor $S^i$ can communicate with nearby sensors within communication range, i.e., $S^j\in\mathcal{N}_t^i$. A target object $G$ is located randomly in the 2D ground plane at position $q^G$. Each sensor predicts its direction $u_t^i \in \mathcal{U}$ along its shortest path towards the target $G$, where $\mathcal{U}$ is a set of $K$ possible discrete directions.
The mobile robot $R$ is located at position $q_t^R$ and moves in the ground plane in $\mathcal{W}\setminus\mathcal{C}$. It is equipped with \textit{any one} of the sensors (we choose sensor $S^1$) and uses the directional output $u_t^1$ of that sensor to execute an action $a_t \in \mathcal{A}$ by applying a velocity of the same direction. \textit{The robot's objective is to move to the target $G$ along the shortest collision-free path. It utilizes the information shared through the sensor network to make an informed decision about how to reach the (potentially occluded) target, while avoiding time-consuming exploration.}

We formalize this as a sequential decision making problem under uncertainty about the underlying world~\cite{bhardwaj2017learning}, and define a corresponding Markov Decision Process (MDP).
At time step $t$, let $s_t \in \mathcal{O}$ be the observed state of the environment, i.e $s_t = \{ o_t^1, \dots, o_t^N \}$. On executing $a_t$, the new state $s_{t+1}$ is determined by the underlying world $\mathcal{W}$, which is a hidden variable, sampled from a prior $P(\mathcal{W})$ and in turn induces a state transition distribution $P(s_{t+1} | s_t, a_t)$. The one-step cost $c(s_t, a_t)$ is the distance travelled since the previous time step and a consequence of action $a_t$.

Let $\pi(s_t)$ be a policy that maps the state $s_t$ to an action $a_t$. The policy represents the navigation strategy that we wish to learn. An episode continues until either the goal is reached ($\lVert q^G - q^R_t \rVert < D_G$) or time horizon $T$ is reached.
Given a prior distribution over worlds $P(\mathcal{W})$ and a distribution over start and goal positions $P(q_0^R, q^G)$, we can estimate the cost of moving from position $q^R_0$ to $q^G$ as
\begin{equation}
    \label{value_function}
    V\left(s_t\right) = \sum_{t=1}^{T}\expect{s_t\sim d^t_\pi}{c(s_t, \pi(s_t))}
\end{equation}
where $d^t_\pi=P(s_t|\pi,\mathcal{W},q_t, q^G)$ is the distribution over states induced by running $\pi$ on the problem $(\mathcal{W},q_t, q^G)$ for $T$ steps \cite{ross2014reinforcement}, and we evaluate the performance of a policy as
\begin{equation}
	\label{eq:policy_cost}
	J\left( \pi \right) = \expect{
  \substack{\mathcal{W} \sim P(\mathcal{W}), \\
  \left(q_0^R, q^G \right) \sim P(q_0^R, q^G)}
  }{V(s_t)}.
\end{equation}

\section{Visual Navigation using Sensor Networks}

\label{sec:visual_nav_method}
This section describes our approach to training the navigation policy $\pi$. \autoref{fig:system} shows the simulation environment and \autoref{fig:nn_architecture} shows an overview of our architecture. The objective of each sensor $S^i$ is to predict a direction $u_t^i$ along the shortest path to the target (with the consideration of static obstacles) by using its own observation $o_t^i$ and the messages shared by other sensors. 
We use a variant of Imitation Learning (IL) using expert solvers (i.e., cheap and fast path planning algorithms).

Our method does not rely on any positioning information whatsoever, and as explained in \autoref{sec:problem_formulation}, the neighborhood $\mathcal{N}_t^i$ is defined through sensors within communication range. During training and evaluation in simulation, we have to model this communication range. We assume a disk model, where the neighbor set is defined as $\mathcal{N}_t^i=\{S^j|L(S^i,S^j)\leq D_S\}$, where $L_t(S^i,S^j)$ is the euclidean distance between $S^i$ and $S^j$ at time $t$, and $D_S$ is the communication range.

In \autoref{sec:dataset}, we explain how we generate the dataset $\mathcal{D}_{\mathrm{IL}} = \{(o_0^{i,m}, \hat{A}_0^{i,m})\}_{m=1,i=1}^{M,N}$ consisting of $M$ per-sensor cost-to-go advantage labels $\hat{A}$ and observations $o$. In \autoref{sec:GNN}, we detail the neural network model $\psi \circ \theta \circ \phi(\cdot)$ consisting of the feature extractor $\phi$, the feature aggregator $\theta$ and the post-processor $\psi$ to predict the advantages. The architecture is depicted in \autoref{fig:nn_architecture}. This neural network is optimized with the objective $J_\mathrm{IL}$ as

\begin{equation}
    J_\mathrm{IL}(\psi \circ \theta \circ \phi) = \expect{o, \hat{A} \sim \mathcal{D}_\mathrm{IL}}{
    \lVert
    \psi \circ \theta \circ \phi(o) - \hat{A}
    \rVert^2}.
\end{equation}

While we train our policy in simulation, we take an approach that facilitates sim-to-real transfer (see~\autoref{sec:sim2real}).

\subsection{Data Generation}\label{sec:dataset}

Imitation learning can be treated as supervised learning problem. Since future states are influenced by previous actions in IL, the i.i.d. assumption of supervised learning is invalidated. \textsc{AggreVate} \cite{ross_aggrevate} proposes one possible solution to this problem by iteratively learning cost-to-go estimates for trajectories of data using an online procedure. For the type of navigation policy that we wish to learn, it is trivial to compute cost-to-go estimates, and therefore we treat it as standard supervised learning problem with an i.i.d. assumption for each sample.

Let $Q^i$ be the cost-to-goal for sensor $S^i$. Let the set of possible directions be $\mathcal{U} = \{u^1, \dots, u^K\}$. Let $Q_{uk}^i$ be the cost-to-goal for sensor $S^i$ upon moving towards direction $u_k^i$. We use the Dijkstra's shortest path algorithm on visibility graphs constructed on environment samples to compute cost-to-goal values $Q^i$. To simplify the training procedure and facilitate better generalization, instead of learning absolute cost-to-goal values $Q^i$, we learn a relative cost-advantage vector $A^i = [Q_{a1}^i - Q^i, \dots, Q_{ak}^i - Q^i]^\intercal$ that subtracts the cost from the current state to the goal for each of the $K$ possible directions and thus is scale-invariant (since the scale of the cost-to-go values are not relevant for a sequential decision making problem).

We generate the dataset $\mathcal{D}_{\mathrm{IL}}$ consisting of $M=40,000$ environment samples for $N=7$ sensors.

\subsection{GNN-based Feature Aggregation across Sensor Network} \label{sec:GNN}

\begin{figure*}[!t]
\centering
    \resizebox{\textwidth}{!}{\input{Figs/gnn_architecture/overview}}

\caption{Overview of our model, with communication visualized for agent $i$. We train a decentralized GNN-based policy that takes camera observations $o_t^i$ as input and outputs directions towards the target $u_t^i$ for each node. The CNN $\hat{\phi}$ encodes images into features $z_t^i$ and the GNN $\theta$ further encodes this for multiple communication hops in $L$ layers as $z_t^{i,l}$. Lastly, the post-processing MLP $\psi(\cdot)$ generates cost-to-go advantages $A_t^i$ which are used to sample a direction to the target along the shortest path as $u_t^i$ and eventually to generate action $a_t$ for the robot, which is equipped with sensor $S_1$. Note that all sensor nodes are trained to predict the direction to increase the number of samples observed, but during deployment, only the robot uses this information.}
\label{fig:nn_architecture}
\end{figure*}
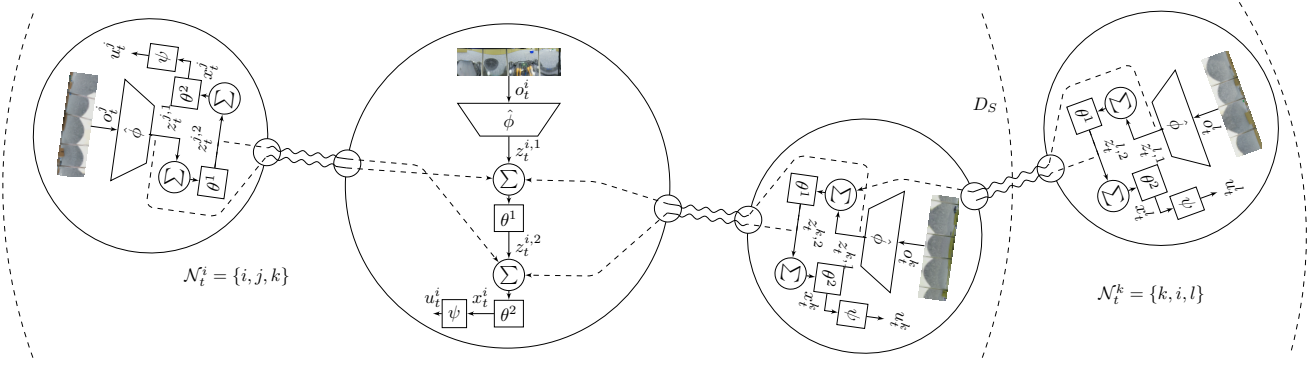

The neural network is homogeneous across all nodes and is divided into the three sub-modules feature extractor $\phi(\cdot)$, feature aggregator $\theta(\cdot)$ and post-processor $\psi(\cdot)$.

\textbf{Local feature extraction.} We first use a Convolutional Neural Network (CNN) $\phi(\cdot)$, specifically MobileNet~v2~\cite{Sandler_2018_CVPR}, which has been pre-trained on ImageNet, to extract features $z_t^i$ from the image $o_t^i$ of each sensor $S^i$. MobileNet is optimized for the evaluation on mobile devices, which makes it an ideal candidate for the application in a distributed sensor network. For the sim-to-real transfer, we replace the encoder, as explained in detail in \autoref{sec:sim2real} and visualized in \autoref{fig:nn_architecture}. The encoding $z_t^i$ has a size of $\mathbb{R}^{32}$ and is communicated to other sensors within communication range.

\textbf{Neighborhood feature aggregation.} In order to predict the target direction, our models need to be able to aggregate information across the whole sensor network. In other words, each sensor requires effective information from those sensors that can \textit{directly} see the target. This feature aggregation task is more challenging than the traditional GNN-based feature aggregation for information prediction \cite{Akter2019} or robot coordination tasks \cite{Li2020, hu2021vgai}. Specifically, in the aforementioned papers, each agent only needs to aggregate information from the nearest few neighbors as their tasks can be achieved by only considering local information. For each agent, information contributed by a very remote agent towards improving the prediction performance can vanish as the network becomes larger. Additionally, in our task, only a limited number of sensors can directly `see' the target. 
{\it Yet, crucially, information about the target from these sensors should be transmitted to the whole network, thus enabling all the sensors to predict the target direction from their own location (which is potentially outside line-of-sight)}). In addition, as we do not introduce any global nor relative pose information, in order to predict the target direction, each sensor must implicitly learn the ability to estimate the relative pose to its neighbors by aggregating image features. Furthermore, generating an obstacle-free path in the target direction by only using image features (without knowing the map) is also very challenging.

The GNN model $\theta^{(L)}(\cdot)$ consists of $L$ layers and generates an encoding $z_t^{i,l}$ for each layer $L$. It takes the image encoding generated by the encoder $z_t^{i,1} = z_t^i$ as input to the first layer, aggregates communicated neighbors' features using the neighborhood $\mathcal{N}_t^i$ over multiple layers, and extracts fused features $x_t^i = z_t^{i, L}$ where $z_t^{i, L}$ is the encoding generated by the last GNN layer for each sensor $S^i$ so that $\theta^{(L)}(z_t^i) = x_t^i$.

A GNN model consists of a stack of neural network layers, where each layer aggregates local neighborhood information, i.e., features of neighbors, around each node and then passes this aggregated information on to the next layer. Specifically, our method builds on the GNN layer introduced in \cite{morris2019weisfeiler}.
We use a recursive definition for a multi-layer GNN where each layer $L > 0$ computes new features as 
\begin{equation}\label{eq:basicgnn}
  \begin{split}
	\theta^{(L)}(z_t^{i,L}) = \sigma \Big(
	& \theta^{(L-1)}(z_t^{i,L}) \cdot  W^{(L)}_1 +\\
	& \sum_{\mathclap{j \in \mathcal{N}_t^i}}\,\, \theta^{(L-1)}(z_t^{j,L}) \cdot W_2^{(L)} \Big)
  \end{split}
\end{equation}
that are communicated in the neighborhood $\mathcal{N}_t^i$ where $W_1^{(L)}$ and $W_2^{(L)}$ are trainable parameter matrices, and $\sigma$ denotes a component-wise non-linear function.

\textbf{Cost-to-goal prediction.} Lastly, we utilize a post-processing Multi-Layer Perceptron (MLP) to predict a set of cost-to-go advantages for each sensor so that $\psi(x_t^i) = A_t^i$, which is eventually used to sample a target direction $u_t^i$ and eventually transform it to an action $a_t$.

\textbf{Policy.} This results in the composition $\psi \circ \theta \circ \phi(o_t^i) = A_t^i$ to compute a cost-to-go advantages from the local image $o_t^i$ and features $z_t^{i,l}$ communicated through the neighborhood $\mathcal{N}_t^i$. We model the set of target directions $\mathcal{U}$ as a discrete distribution so that $u_t^i = \mathrm{cat}(\sigma(-\alpha A_t^i))$ where $\mathrm{cat}(\cdot)$ samples from a categorical distribution, $\sigma(\cdot)$ is the softmax function, and $\alpha$ is a hyperparameter to adjust the stochasticity, which may help the robot to avoid oscillating between two possible actions. As explained in \autoref{sec:problem_formulation}, the relationship between direction $u$ and action $a$ is bijective. Hence, any sensor can be used as part of a mobile robot (to command its motion). In this work, we denote that sensor as $S_1$. So far, we have used a distributed notation to compute a direction $u_t^i$ for any given sensor. We formulate the target navigation problem as centralized MDP relying on a central state $s_t = \{ o_t^1, \dots, o_t^N \}$ consisting of the observation of all sensors. Therefore, we can define the policy $\pi$ as

\begin{align}\label{eq:policy}
	\pi(\{ o_t^1, \dots, o_t^N \}) &= \mathrm{cat}(\sigma(-\alpha (\psi \circ \theta \circ \phi(o_t^1)))) \beta \\
	&= u_t^1 \beta = a_t,
\end{align}

where $\beta$ is a hyperparameter to scale the magnitude and thus transforming a direction into a velocity action $a_t \in \mathcal{A}$. Note that even though the notation of the policy is centralized, the execution is inherently decentralized, and instead of depending on raw image observations for other sensor nodes, the robot only depends on the local observation $o_t^1$ and latent encodings communicated through the GNN $\theta$.


\section{Zero-Shot Real World Transfer}
\label{sec:sim2real}

To demonstrate the feasibility of a zero-shot transfer of our model, we design a twin environment setup, consisting of both real and digital copies of the operational space, see \autoref{fig:system}. The real setup includes custom-built sensors that provide local visual sensing to nearby nodes within the communication network. Later, in \autoref{sec:results}, we report results that demonstrate the effectiveness our sim-to-real approach for the real-world scenario.

\subsection{Setup}

\textbf{Twin Environments}.
The real-world environment is 5.7\,m $\times$ 4.2\,m, and is cluttered with three to five obstacles of different size. We use standard cardboard boxes for obstacles and blue and green building blocks to identify sensor locations and target location, respectively.
In order to facilitate zero-shot transfer, we design a digital twin of our real-world environment. This digital twin (i.e., a simulation environment) is built within the Webots simulator~\cite{Webots04} and is illustrated in \autoref{fig:system}. 

\textbf{Custom-made sensor nodes}. We design and construct six sensor nodes that consist of a contraption holding the downward-facing camera with a fisheye lens as well as a local data processing unit. 

\textbf{Mobile robot}.
We use the DJI RoboMaster as mobile robot platform.
A seventh sensor node that also serves as controller for the robot is mounted on top. 
The robot employs a collision shielding mechanism, 
that takes as input distance measurements to detect the near-sided obstacles and the border of environment. Repulsive force against detected obstacles are generated through a potential field~\cite{hwang1992potential}. In practice, during real-world deployment, we have a fully connected communication network due to wireless communication ranges exceeding the size of our indoor environment.

\subsection{Domain Adaptation}

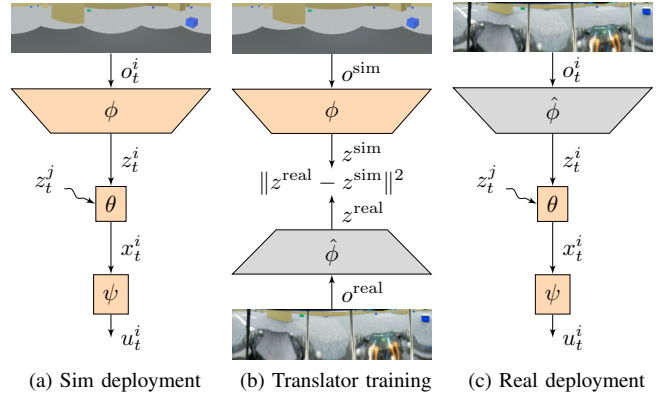
\begin{figure}[!t]
\subfloat[Sim deployment]{\resizebox{0.32\linewidth}{!}{\input{Figs/sim_to_real/sim_deployment}}}
\hfill
\subfloat[Translator training]{\resizebox{0.32\linewidth}{!}{\input{Figs/sim_to_real/translator}}}
\hfill
\subfloat[Real deployment]{\resizebox{0.32\linewidth}{!}{\input{Figs/sim_to_real/real_world_deployment}}\label{fig:env_real_stage_2}}

\caption{Our sim-to-real framework.
    (a) We first train and deploy the policy in simulation.
    (b) We proceed to collect image pairs from simulation $o^\mathrm{sim}$ and the real world $o^\mathrm{real}$ and use them to train a real-to-sim translator model $\hat{\phi}$ (gray) by reconstructing latent features from the simulation domain $z$ generated through the encoder trained in simulation $\phi$ (orange) with real-world images by minimizing the loss function $J_\mathrm{s2r}$.
    (c) We combine the translator model $\hat{\phi}$ trained on real-world images (gray) with $\theta$ and $\psi$ trained in simulation (orange) to deploy the policy to a real-world setup.}
    \label{fig:nn_sim_to_real}
\end{figure}

As outlined in \autoref{fig:nn_architecture}, we first train the policy using simulated images through IL. \autoref{fig:nn_sim_to_real} outlines how we perform the sim-to-real transfer. We create a real-world environment and map it using a motion capture system. We transfer this map into Webots to have an identical representation of the environment in the real world and in simulation. To train the translator, we tele-operate the robot in this environment to collect $M$ image pairs of real-world images $o_\mathrm{real}$ and corresponding simulated images $o_\mathrm{sim}$ and store them in a dataset $\mathcal{D}_\mathrm{s2r} = \{(o_\mathrm{real}^j, o_\mathrm{sim}^j)\}_{m=0}^M$.

In total, we construct 8 different environments, each populated with 6 sensors and one robot. We collect 2000 image pairs for each sensor in each environment (this is achieved within 5 minutes with images being recorded at 10 Hz). Constructing and mapping each environment and setting up the data collection procedure takes 30 minutes per environment. We automatically filter the images in post processing for the sensor images to only be included when the robot is moving within the field of view of the sensors. This results in a dataset of $M=80,000$ image pairs for the training set and $10,000$ images for the test set.

We use a similar approach to \cite{zhang2019vr}, where a neural network based translator model is trained to map real-world images to simulated images, which are then fed to the policy, but instead of mapping to simulated images, we map to their respective encoding. Specifically, we train a translator model $\hat{\phi}(\cdot)$ that maps real images $o^\mathrm{real}$ to the encoding of the corresponding simulated image $\phi(o^\mathrm{sim})$ by minimizing the objective $J_\mathrm{s2r}(\cdot)$ while keeping $\phi(\cdot)$ fixed,
\begin{equation}
    J_\mathrm{s2r}(\hat{\phi}) = \expect{o^\mathrm{sim}, o^\mathrm{real} \sim \mathcal{D}_\mathrm{s2r}}{\lVert \hat{\phi}(o^\mathrm{real}) - \phi(o^\mathrm{sim}) \rVert^2}.
\end{equation}

\section{Results}
\label{sec:results}

We first introduce the metrics we use for evaluation and then demonstrate the performance of our method in simulation, based on the methodology introduced in \autoref{sec:visual_nav_method}. Second, we demonstrate successful zero-shot transfer to the real-world setup introduced in \autoref{sec:sim2real}. 

\subsection{Performance Metrics}\label{sec.Met}

We evaluate the trained policies on the unseen test split of the training dataset as well as a generalization set that has been generated with a larger environment and more sensors.

\begin{figure*}[bt]
\centering
\subfloat[Env. A]{\includegraphics[height=3.5cm]{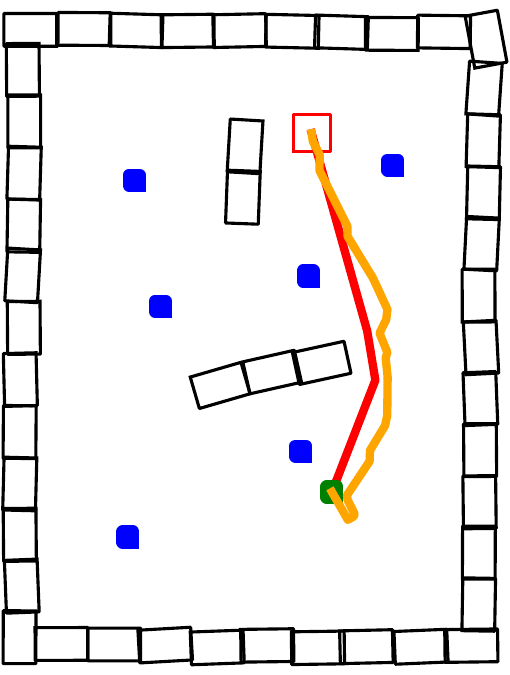}
\includegraphics[height=3.5cm]{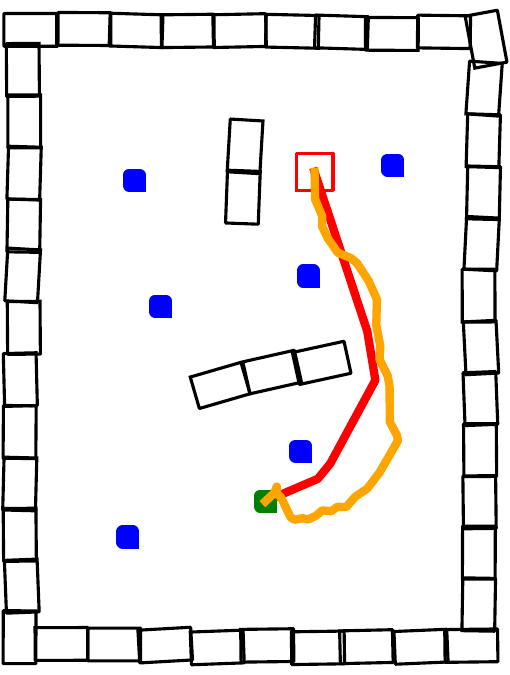}
}\hfill
\subfloat[Env. B]{\includegraphics[height=3.5cm]{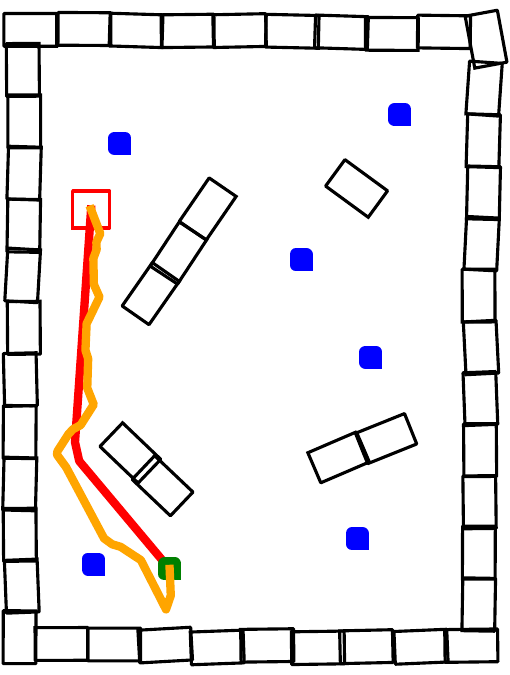}
\includegraphics[height=3.5cm]{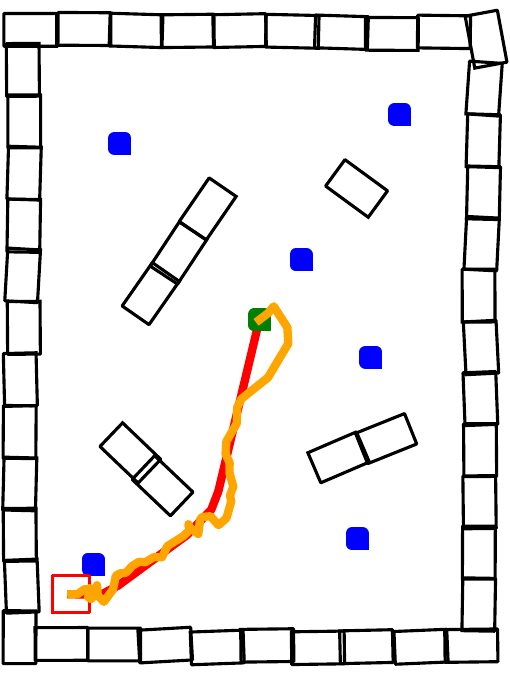}
}\hfill
\subfloat[Env. C]{\includegraphics[height=3.5cm]{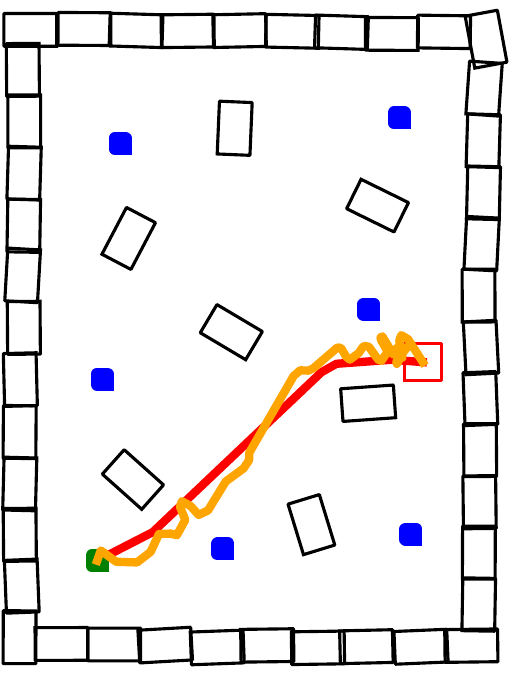}
\includegraphics[height=3.5cm]{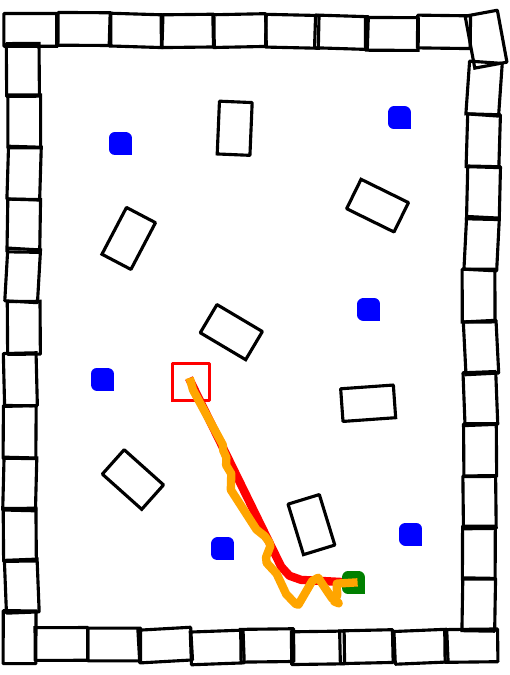}
}
\caption{A selection of policy evaluations for each real-world environment. Blue squares indicate sensor positions, green squares the target position $q^G$, red squares the robot's initial position $q_0^R$, red paths the shortest path computed by the expert $P$ and the orange path the path $p$ chosen by the policy $\pi$.}
\label{fig:real_evaluation}
\vspace{-0.2cm}
\end{figure*}

We consider two primary metrics for our evaluation. A run is considered successful if the robot arrives at the target without collisions and within a pre-defined time horizon $T$, captured by the boolean success indicator $C_m$ for case $m$. The \textit{success rate} is the fraction of all successful runs as $\frac{1}{M}\sum_{m=1}^{M} C_m$. We furthermore report the Success Weighted by Path Length, or SPL \cite{anderson2018evaluation}, as $\frac{1}{M}\sum_{m=1}^{M} C_m \frac{P_m}{\mathrm{max}(p_m,P_m)}$, where $P_m$ is the shortest path length from the robot's initial position to the target, and $p_m$ the length of the path actually taken. We visualize a selection of paths in \autoref{fig:real_evaluation}. We report all results for environment configurations where the target is outside line-of-sight, therefore requiring additional sensor coverage for an efficient solution.

\subsection{Simulations}

We train five variants of our policy to evaluate our approach. We first train a policy with communication range $D_S=0.0\:\mathrm{m}$ as a baseline (i.e., no communication). We also train three policies with communication ranges of $D_S=2.0\:\mathrm{m}$, $D_S=4.0\:\mathrm{m}$ and $D_S=\infty\:\mathrm{m}$ (fully connected) respectively, all with a single GNN layer. Lastly, we train a policy with a communication range of $D_S=2.0\:\mathrm{m}$ and two GNN layers. The results can be seen in \autoref{tab:sim_results}.

We evaluate all policies on unseen maps on the small test set (environment sized $W=8\:\mathrm{m}$ and $H=10\:\mathrm{m}$ with $N=7$ sensors, $M=52$ trials) and the large generalization set (environment sized $W=16\:\mathrm{m}$ and $H=20\:\mathrm{m}$ with $N=13$ sensors, $M=42$ trials).
The baseline policy without communication has a success rate of $0.827$ and an SPL of $0.719$ in the small environment. These values are the lower performance bound for no communication. We ensure that all environments are solvable. Even without communication, the target can be discovered through random exploration. The success metrics increase over all experiments with increasing communication ranges, up to $1.0$ success and $0.925$ SPL for the policy trained with $D_S=2\:\mathrm{m}$ and $L=1$. The policies with $D_S=2.0\:\mathrm{m}$ and $L=2$ perform similar to the policy with $D_S=4\:\mathrm{m}$. The policy with $D_S=\infty\:\mathrm{m}$ and $L=1$ performs slightly worse. This can be attributed to the unfiltered inclusion of information from all sensors. Adding locality through a neighborhood and considering multiple neighborhood through multi-hop communication helps in building a more appropriate global representation.

\begin{figure*}[bt]
\centering
\includegraphics[height=3.5cm]{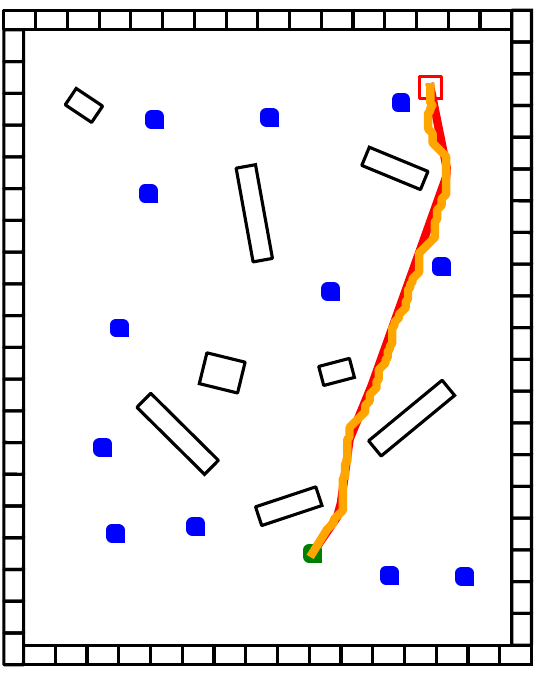}
\includegraphics[height=3.5cm]{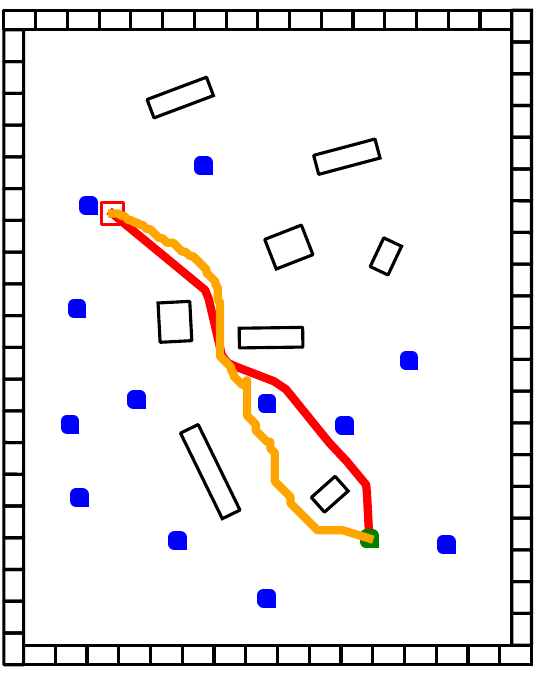}
\includegraphics[height=3.5cm]{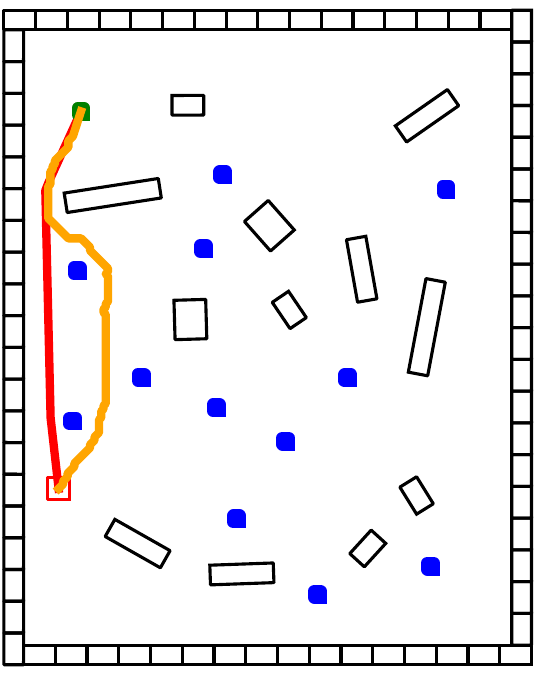}
\includegraphics[height=3.5cm]{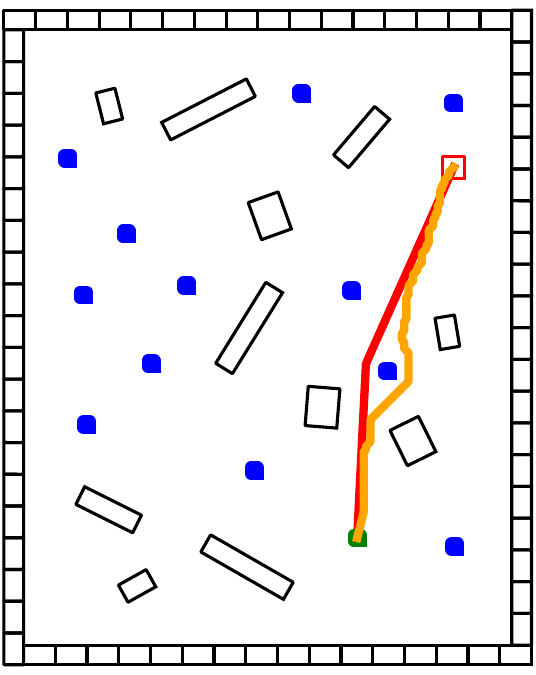}
\includegraphics[height=3.5cm]{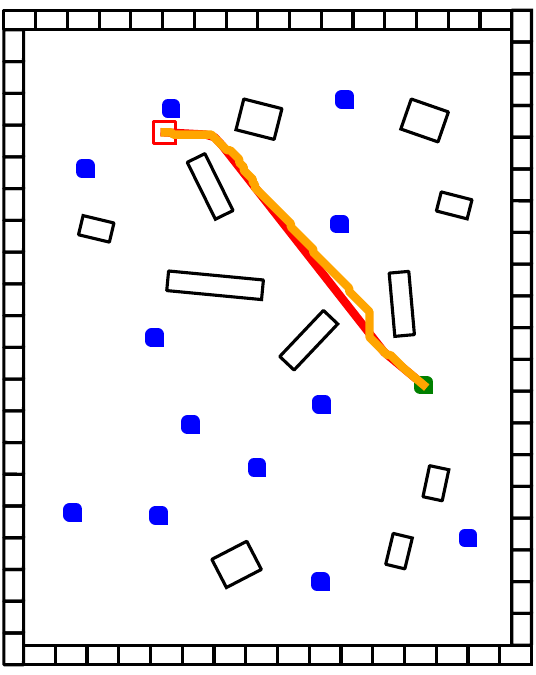}
\includegraphics[height=3.5cm]{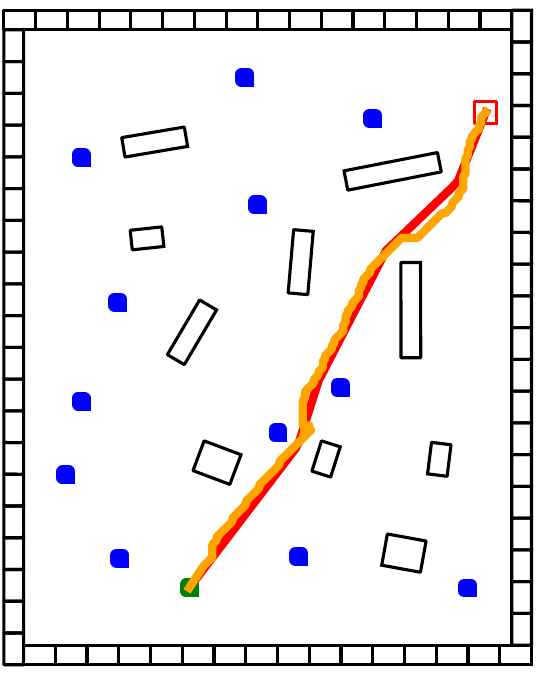}
\caption{{A selection of policy evaluations for a policy trained on a small environment ($W=8\:\mathrm{m}$, $H=10\:\mathrm{m}$, $N=7$) and evaluated on a large environment ($W=16\:\mathrm{m}$, $H=20\:\mathrm{m}$, $N=13$). Blue squares indicate sensor positions, the green square the target position $q^G$, the red path the robot's initial position $q_0^R$, the red path the shortest path computed by the expert $P$ and the orange path the path $p$ chosen by the policy $\pi$. The policy generalizes.}}
\label{fig:sim_evaluation}
\vspace{-0.2cm}
\end{figure*}

We furthermore test the generalizability to larger environments. The baseline policy has a success rate of $0.619$ and an SPL of $0.492$, while the fully connected policy has a success rate of up to $0.952$ and an SPL of $0.853$. Samples from this evaluation are visualized in \autoref{fig:sim_evaluation}.

\begin{figure}[bt]
\centering
\includegraphics{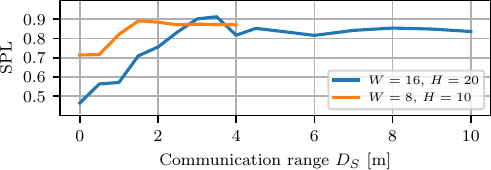}\hfill
\vspace{0.2cm}
\includegraphics{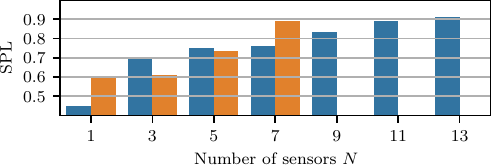}
\caption{We record the performance of the policy trained for $D_S=2\:\mathrm{m}$ and $L=2$ for a range of different communication ranges and number of sensors. Top: The large environment (blue) is populated with $N=13$ sensors and the small environment (orange) with $N=7$ sensors. Bottom: The large environment (blue) has a communication range of $D_S=3.5\:\mathrm{m}$ and the small environment (orange) of $D_S=1.5\:\mathrm{m}$. Both values result in peak performance for the maximum number of sensors in each environment. Increasing the communication range and number of sensors benefits the SPL.}
\label{fig:eval_comm_range}
\end{figure}

In \autoref{fig:eval_comm_range} we further analyze the benefit of communication on our method. We use the policy trained for $D_S=2\:\mathrm{m}$ and $L=2$ on the small and the large test set and evaluate the performance for a variety of communication ranges and number of sensors. Note that the results in \autoref{tab:sim_results} for the large environment are suboptimal due to the constrained communication range, since $D_S=2.0\:\mathrm{m}$ for $L=2$ covers sensors at most $4.0\:\mathrm{m}$ away, while the maximum environment length is $14.0\:\mathrm{m}$. We find that the SPL in the large environment increases approximately linearly from $0.45$ for $D_S=0.0\:\mathrm{m}$ to $0.91$ for $D_S=3.5\:\mathrm{m}$ and then slightly decreases to a constant of $0.85$ for $D_S \ge 4.0\:\mathrm{m}$, resulting in a performance increase of $2.0\times$ compared to the communication-free baseline. The small environment performs at $0.72$ SPL without communication and $0.93$ with communication, resulting in a $1.3\times$ performance increase. Decreasing the number of sensors correspondingly decreases the SPL for a similar minimum and maximum performance. It is to be expected that both environments perform similarly well with the maximum number of sensors and a communication range that results in a coverage of the whole environment, whereas in the no communication case and with only one sensor, the small environment performs better (since less exploration is required to navigate to the target).

\begin{table}[t]
\vspace{0.2cm}
\small
\centering
\caption{Evaluation of sim result for policies trained for different communication ranges $D_S$ and number of GNN layers $L$.
}
\label{tab:sim_results}

\begin{tabular}{cc|cc|cc}
      \toprule
      &     & \multicolumn{2}{c|}{\begin{tabular}[c]{@{}c@{}}Training Distribution\\ \end{tabular}} & \multicolumn{2}{c}{\begin{tabular}[c]{@{}c@{}}Generalization \end{tabular}} \\
$D_S$ & $L$ & Success                 & SPL                 & Success                 & SPL                \\
      \midrule

0.0 & 1 & $0.827$ & $0.719$ & $0.619$ & $0.492$ \\
\midrule
2.0 & 1 & $1.000$ & $0.925$ & $0.833$ & $0.742$ \\
4.0 & 1 & $0.981$ & $0.912$ & $0.905$ & $0.804$ \\
2.0 & 2 & $0.962$ & $0.885$ & $0.857$ & $0.756$ \\
$\infty$ & 1 & $0.962$ & $0.880$ & $0.952$ & $0.853$ \\

      \bottomrule
\end{tabular}
\end{table}

\subsection{Sim-to-Real Policy Transfer}

After performing the domain adaptation, we evaluate the performance of the policy on the real-world setup as described in \autoref{sec:sim2real}. We randomly construct three different environments and run between 19 and 29 runs per environment, resulting in a total of 69 evaluation runs.

\begin{figure}[bt]
\centering
\subfloat[Failure]{\includegraphics[height=2.6cm]{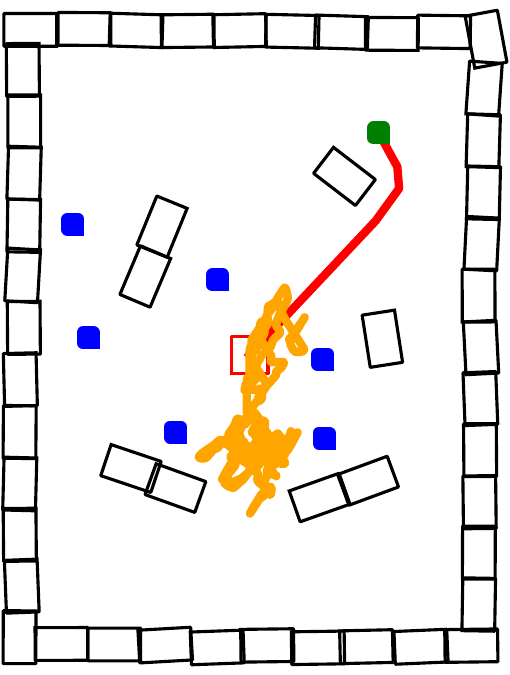}
}
\subfloat[Success]{\includegraphics[height=2.6cm]{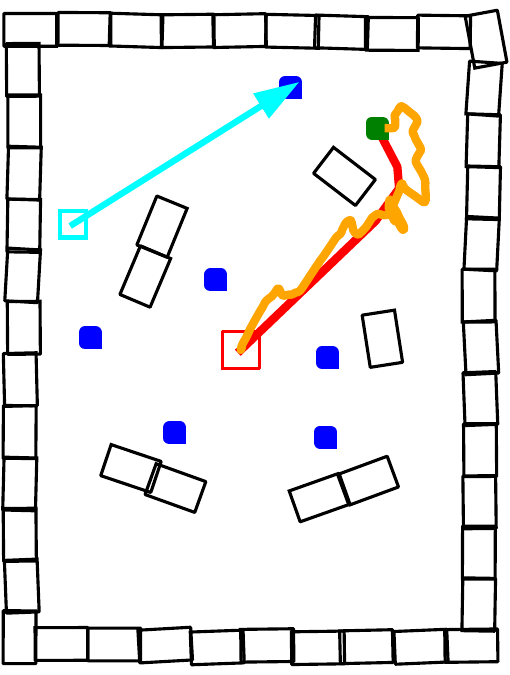}
}
\subfloat[Success]{\includegraphics[height=2.6cm]{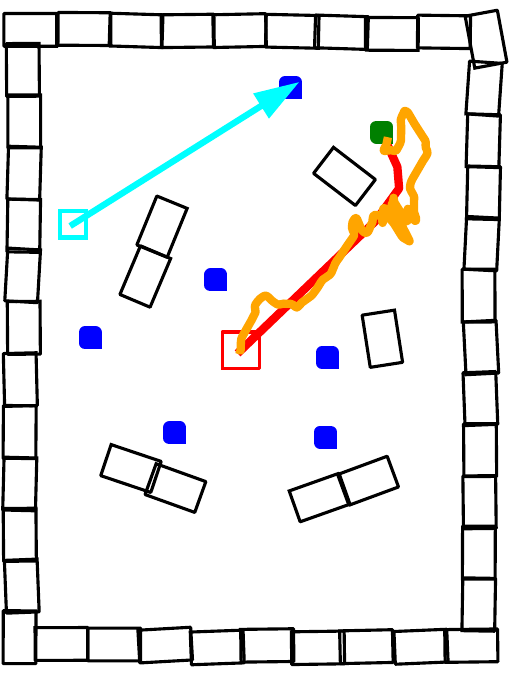}
}
\subfloat[Success]{\includegraphics[height=2.6cm]{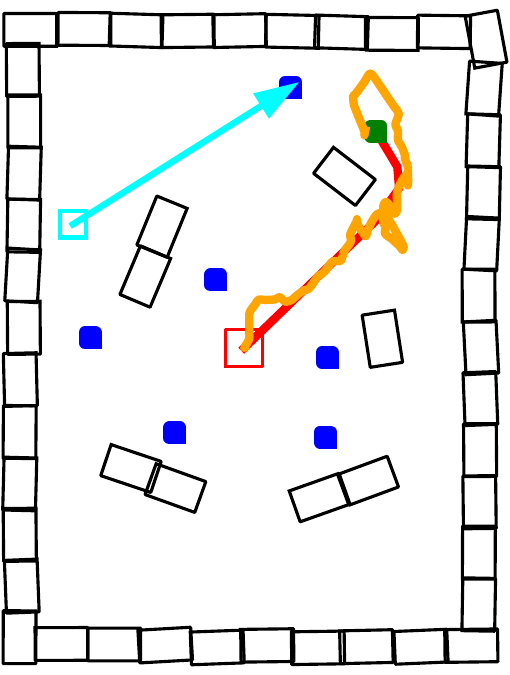}
}

\subfloat[Failure]{\includegraphics[height=2.6cm]{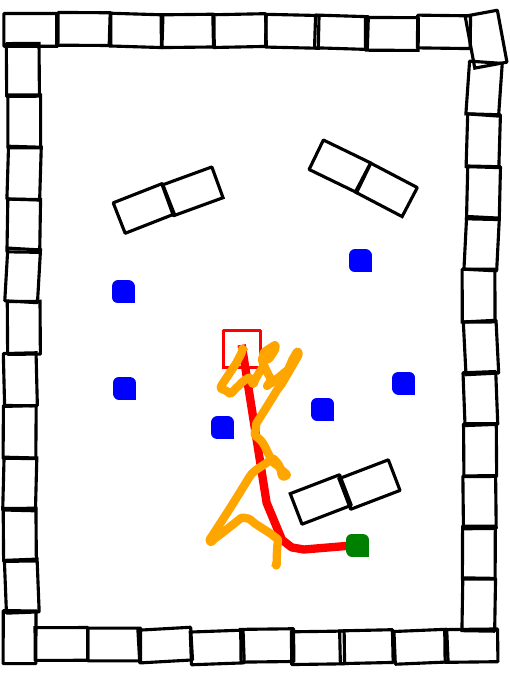}
}
\subfloat[Success]{\includegraphics[height=2.6cm]{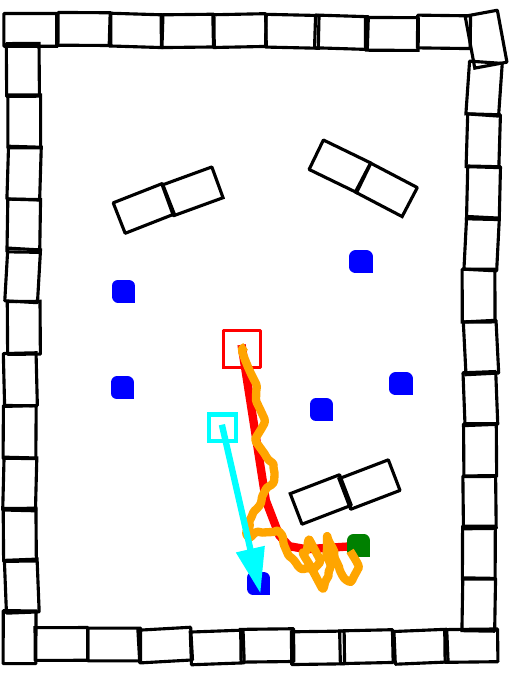}
}
\subfloat[Success]{\includegraphics[height=2.6cm]{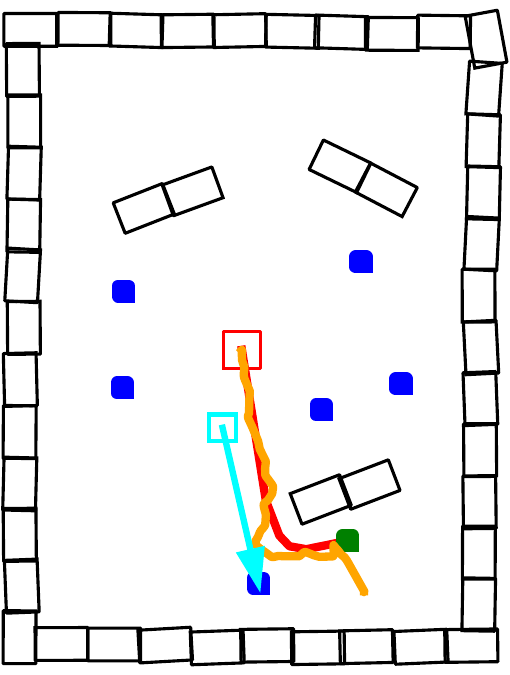}
}
\subfloat[Success]{\includegraphics[height=2.6cm]{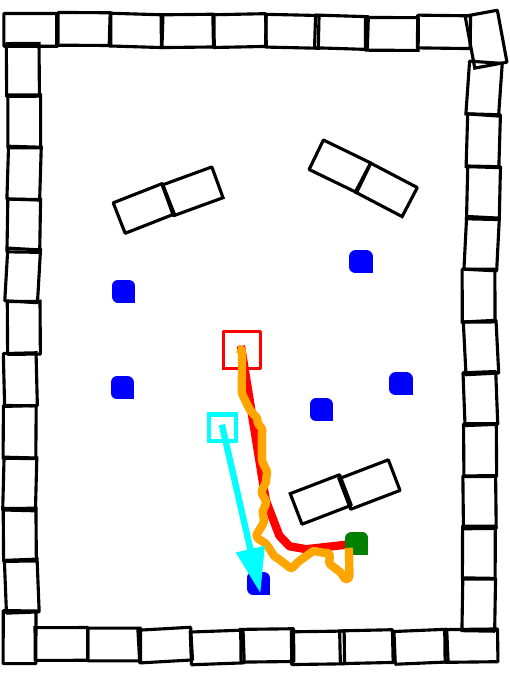}
}

\subfloat[Failure]{\includegraphics[height=2.45cm]{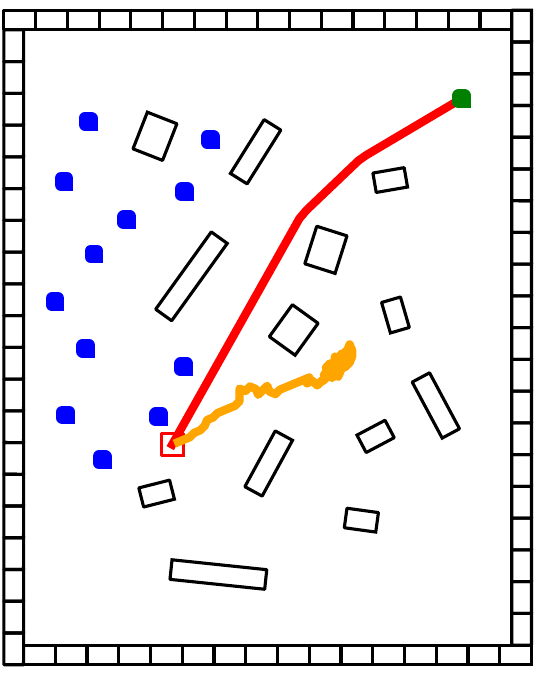}
}
\subfloat[Success]{\includegraphics[height=2.45cm]{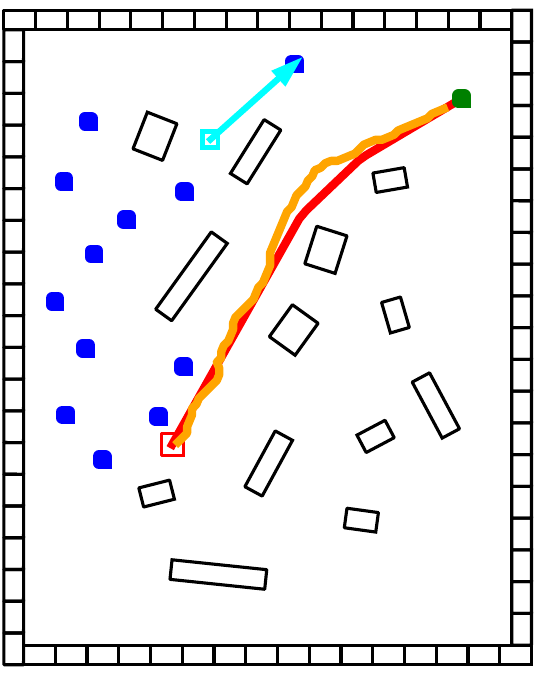}
}
\subfloat[Success]{\includegraphics[height=2.45cm]{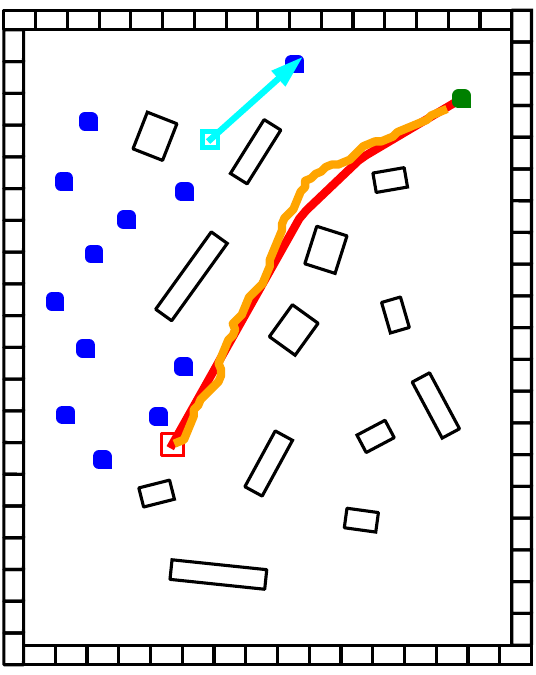}
}
\subfloat[Success]{\includegraphics[height=2.45cm]{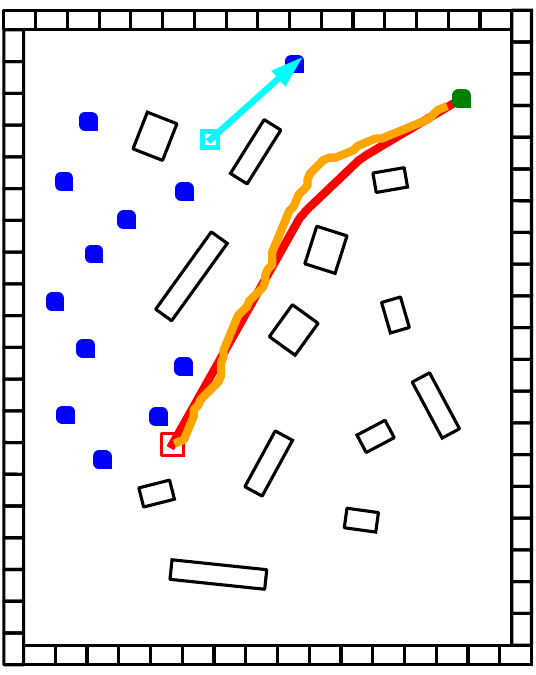}
}
\caption{{We showcase the capability of our method to adapt to changing sensor layouts. We first construct a real-world environment in which no sensor has visual coverage of the target, (a), (e), and (i). The robot is not able to locate the target. We then move a sensor within line-of-sight to the target (cyan). The robot is now able to infer its direction w.r.t. the target and is, therefore, able to navigate towards it ((b-d), (f-h) and (j-l)). The first two rows are performed in the real-world setup, and the last row in simulation.}}
\label{fig:real_evaluation_occlusion}
\end{figure}

We first perform experiments to showcase the ability of the policy to dynamically respond to changes in sensor positions, which is useful in cases where the sensors are mounted on mobile platforms  \autoref{fig:real_evaluation_occlusion}. The trials show that the robot is not able to locate the target if it is not within line-of-sight of at least one sensor (see the (a), (e) and (i)). As soon as one sensor is moved within line-of-sight to the target, the robot is able to successfully navigate to the target. Our experiments furthermore indicate that our approach is able to handle dynamically changing obstacle positions, as long as they are observed by a sensor.

The results for all environments are shown in \autoref{fig:real_results}. The total success rate across all environments is $0.957$ and $0.574$ SPL. Environment A is the least challenging environment, with only two obstacles in total, and has the highest success rate of $1.0$ and SPL of $0.588$. Environment C is, with a success rate of $0.905$, the most challenging one, with a total of seven obstacles, ranging from small to large, and narrow passages between obstacles. Environment B has more obstacles than Environent B but less than Environment C with a success rate of $0.947$ and an SPL of $0.634$. Even though the sim-to-real transfer is generally successful, there is a noticeable reality gap. This can be attributed to the physics reality gap, which has not been fully addressed in this work, and causes oscillations when moving in the real-world. Even though we did not conduct any real-world baseline experiments, it is to be expected that the SPL and success rate for $D_S = 0 $ (i.e., no communication) is significantly lower, as observed in our simulation results.

\begin{table}[bt]
\vspace{0.2cm}
\centering
\small
\caption{Evaluation of real-world results for three different environments for the fully connected policy. We report the number of trials $M$ for each environment.}
\label{fig:real_results}

\begin{tabular}{c|ccc}
    \toprule
Env & $M$ & Success       & SPL     \\
\midrule

A & 29 & $1.000$ & $0.588$ \\
B & 19 & $0.947$ & $0.634$ \\
C & 21 & $0.905$ & $0.502$ \\
\midrule
All & 69 & $0.957$ & $0.574$ \\

\bottomrule
\end{tabular}
\end{table}

\section{Conclusions}
We propose a vision-only-based learning approach that leverages a Graph Neural Network (GNN) to encode and communicate relevant viewpoint information to the mobile robot. 
In our experiments, we first demonstrate generalization to previously unseen environments with various sensor network layouts. 
Our results show that by using communication between the sensors and the robot, we achieve a $1.3\times$ improvement on small environments and a $2.0\times$ improvement in SPL on large environments, when compared to the communication-free baseline, hence showing increasing improvement for larger environment sizes. This is done without requiring a global map, positioning data, nor pre-calibration of the sensor network, which we demonstrate in real-world experiments. The benefit of utilizing communication in wireless sensor networks increases as the size of the environment increases, since the robot is less likely to discover the goal through random (unguided) exploration using the communication-free baseline.
We perform a zero-shot transfer from simulation to the real world. To this end, we train a translator model that translates between real and simulated images so that the navigation policy (which is trained entirely in simulation) can be used directly on the real robot, without additional fine-tuning. 
Physical experiments demonstrate first-of-a-kind results that show successful real-world demonstrations on a practical robotic platform with raw visual inputs. In future work, we will evaluate the impact of asynchronous and delayed communication between sensors and the robot and analyze the impact of small vs large visual overlaps (among sensor nodes) on the overall navigation performance.

\section*{Acknowledgement}
This work was supported by European Research Council (ERC) Project 949949 (gAIa). J. Blumenkamp acknowledges the support of the 'Studienstiftung des deutschen Volkes' and an EPSRC tuition fee grant. This research is funded in part by a gift from Arm. Their support is gratefully acknowledged.

%% file: Figs/gnn_architecture/overview.tex
\begin{tikzpicture}[auto,>=latex']
    \def\centerarc[#1](#2)(#3:#4:#5)
        { \draw[#1] ($(#2)+({#5*cos(#3)},{#5*sin(#3)})$) arc (#3:#4:#5); }
    \tikzset{
        enc/.style = {trapezium, trapezium angle=-75, draw, trapezium stretches=true},
        wheel/.style = {draw=black, fill=lightgray},
        pics/enc/.style args={#1, #2}{code={

            \node [coordinate] (_origin) {};

            \node [enc, minimum width=2cm, above left=0em and -1em of _origin, transform shape] (_enc) {$\hat{\phi}$};
            \node[inner sep=0pt, above=1.5em of _enc] (_img) {\includegraphics[width=2cm]{#2}};



            \draw[->](_img.south) -- node[pos=0.5,right] {$o_t^#1$} (_enc.north);
        }},
        pics/wireless/.style args={}{code={
            \node [circle, fill=white, inner sep=0.1em, draw, minimum width=1.5em] (_comm) {};
            \node [coordinate, left=0.25em of _comm.center] (_l) {};
            \node [coordinate, right=0.25em of _comm.center] (_r) {};
        }},
        pics/sensor/.style args={#1, #2}{code={
            \pic[] (_enc) {enc={#1, #2}};

            \node [circle, inner sep=0.1em, draw, below left=1em and 0.6em of _enc_enc] (_agg_1) {$\sum$};
            \node [draw=black, below=0.5em of _agg_1] (_gnn_1) {$\theta^1$};

            \node [circle, inner sep=0.1em, draw, right=3em of _gnn_1] (_agg_2) {$\sum$};
            \node [draw=black, above=0.5em of _agg_2] (_gnn_2) {$\theta^2$};
            \node [draw=black, above right=0em and 0.25em of _gnn_2] (_psi) {$\psi$};

            \node [coordinate, above=1em of _psi] (_action) {};

            \node[circle, draw, minimum size=13em, inner sep=0, transform shape] at (_enc_origin) (_body) {};

            \pic[below=5.75em of _enc_origin] (_tx) {wireless={}};
            \pic[below=10.25em of _enc_origin] (_rx) {wireless={}};

            \draw[->](_enc_enc.south) |- node[pos=0.3,right] {$z_t^{#1, 1}$} (_agg_1.east);
            \draw[-, dashed, rounded corners](_enc_enc.south) -- ++(0em, -0.3em) -- ++(-4em, 0em) -- ++(0em, -4em) -- (_tx_l);
            \draw[-, dashed](_gnn_1.east) -| (_tx_r);
            \draw[->](_agg_1.south) -- (_gnn_1.north);
            \draw[->](_gnn_1.east) -- node[pos=0.5,above] {$z_t^{#1, 2}$} (_agg_2.west);
            \draw[->](_agg_2.north) -- (_gnn_2.south);
            \draw[->](_gnn_2.east) -| node[pos=0.3,below] {$x_t^#1$} (_psi.south);
            \draw[->](_psi.north) -| node[pos=1.0,above] {$u_t^#1$} (_action);

            \draw[-, comm](_tx_l) -- (_rx_l);
            \draw[-, comm](_tx_r) -- (_rx_r);
        }},
        pics/robot/.style args={#1}{code={
            \node[circle, draw, minimum size=#1*1em, inner sep=0] (_body) {};
            \node[coordinate] (_origin) {};

            \def\wheeldist{0.35em}
            \def\wheelbase{0.2em}
            \def\wheelheight{0.125em}
            \def\wheelwidth{0.25em}
            \draw[wheel] ++(#1*\wheeldist,#1*\wheelbase) rectangle ++(#1*\wheelheight,#1*\wheelwidth);
            \draw[wheel] ++(#1*-\wheeldist,#1*\wheelbase) rectangle ++(-#1*\wheelheight,#1*\wheelwidth);
            \draw[wheel] ++(#1*\wheeldist,#1*-\wheelbase) rectangle ++(#1*\wheelheight,-#1*\wheelwidth);
            \draw[wheel] ++(-#1*\wheeldist,-#1*\wheelbase) rectangle ++(-#1*\wheelheight,-#1*\wheelwidth);
        }},
        pics/agent/.style args={#1}{code={
            \pic[yshift=-0.3em] (_enc) {enc=#1};
        }},
        relative to node/.style={
            shift={(#1.center)},
            x={(#1.east)},
            y={(#1.north)},
        },
        comm/.style = {decorate,decoration={snake, amplitude=0.4mm, segment length=3mm, post length=1mm}},
    }
    \node[coordinate] (origin) {};

    \pic[yshift=2.75em] (enc_i) at (origin) {enc={i, {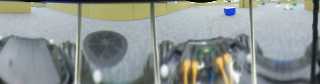}}};
    \node[circle, draw, minimum size=18em, inner sep=0, transform shape] at (origin) (_body) {};
    \node [circle, inner sep=0.1em, draw, below=1.5em of enc_i_enc] (enc_i_agg_1) {$\sum$};
    \node [draw=black, below=0.5em of enc_i_agg_1] (enc_i_gnn_1) {$\theta^1$};
    \node [circle, inner sep=0.1em, draw, below=1.5em of enc_i_gnn_1] (enc_i_agg_2) {$\sum$};
    \node [draw=black, below=0.5em of enc_i_agg_2] (enc_i_gnn_2) {$\theta^2$};
    \node [draw=black, left=1.5em of enc_i_gnn_2] (enc_i_psi) {$\psi$};
    \node [coordinate, left=0.5em of enc_i_psi] (enc_i_action) {};

    \draw[->](enc_i_enc.south) -- node[pos=0.5,right] {$z_t^{i, 1}$} (enc_i_agg_1.north);
    \draw[->](enc_i_agg_1.south) -- (enc_i_gnn_1.north);
    \draw[->](enc_i_gnn_1.south) -- node[pos=0.5,right] {$z_t^{i, 2}$} (enc_i_agg_2.north);
    \draw[->](enc_i_agg_2.south) -- (enc_i_gnn_2.north);
    \draw[->](enc_i_gnn_2.west) -- node[pos=0.5,above] {$x_t^i$} (enc_i_psi.east);
    \draw[->](enc_i_psi.west) -- node[pos=1.0,above] {$u_t^i$} (enc_i_action);
    
    \foreach \angle / \agent / \imgpath in {172/j/{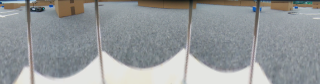}, -8/k/{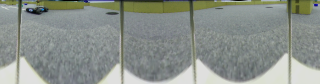}} {
        \pic[rotate fit=\angle + 270, above right=cos(\angle-90)*20em and sin(\angle+90)*20em of origin, anchor=center, transform shape] (agent_\agent) {sensor={\agent, \imgpath}};
    }

    \foreach \angle/\agent/\imgpath in {20/l/{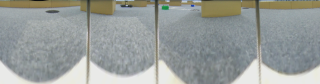}} {
        \pic[rotate fit=\angle + 270, above right=cos(\angle-90)*17.5em and sin(\angle+90)*17.5em of agent_k_enc_origin, anchor=center, transform shape] (agent_\agent) {sensor={\agent, \imgpath}};
    }
    \draw[->, dashed, rounded corners](agent_l_rx_l) -- ++(-4em, 0.75em) -- (agent_k_agg_1);
    \draw[->, dashed, rounded corners](agent_k_rx_l) -- ++(-4em, 1.25em) -- (enc_i_agg_1);
    \draw[->, dashed, rounded corners](agent_k_rx_r) -- ++(-4em, -3.5em) -- (enc_i_agg_2);
    \draw[->, dashed, rounded corners](agent_j_rx_l) -- (enc_i_agg_1);
    \draw[->, dashed, rounded corners](agent_j_rx_r) -- ++(4em, -0.35em) -- (enc_i_agg_2);

    \def\commrange{28em}
    \def\commrangeangle{20}
    \centerarc[dashed](_body)(180-\commrangeangle:180+\commrangeangle:\commrange);
    \centerarc[dashed](_body)(\commrangeangle:-\commrangeangle:\commrange);
    
    \node[above right=4.5em and 26.5em of origin, anchor=center] {$D_S$};
    \node[below left=5em and 15em of origin, anchor=center] {$\mathcal{N}_t^i = \{i, j, k\}$};

    \centerarc[dashed](agent_k_body)(32:-15:24.5em);
    \node[above right=2em and 12em of agent_k_body, anchor=center] {$\mathcal{N}_t^k = \{k, i, l\}$};

\end{tikzpicture}

%% file: Figs/sim_to_real/sim_deployment.tex
\begin{tikzpicture}[auto,>=latex']
    \input{Figs/sim_to_real/tikzset}
    \node[inner sep=0pt] (enc_i_img) {\includegraphics[width=3cm]{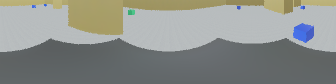}};
    \node [enc, sim, below=1.5em of enc_i_img, transform shape] (enc_i_enc) {$\phi$};
    \node [below left=1.0em and 2em of enc_i_enc.south] (node_j_msg) {$z_t^j$};
    \node [block, sim, below=2.25em of enc_i_enc] (enc_i_gnn) {$\theta$};
    \node [block, sim, below=2.25em of enc_i_gnn] (enc_i_psi) {$\psi$};
    \node [coordinate, below=1.0em of enc_i_psi] (enc_i_action) {};

    \draw[->](enc_i_img.south) -- node[node_label, pos=0.5,right] {$o_t^i$} (enc_i_enc.north);
    \draw[->](enc_i_enc.south) -- node[node_label, pos=0.5,right] {$z_t^i$} (enc_i_gnn.north);
    \draw[comm, ->](node_j_msg) -- (enc_i_gnn.west);
    \draw[->](enc_i_gnn.south) -- node[node_label, pos=0.5,right] {$x_t^i$} (enc_i_psi.north);
    \draw[->](enc_i_psi.south) -- node[node_label, pos=1.0,right] {$u_t^i$} (enc_i_action);

\end{tikzpicture}

%% file: Figs/sim_to_real/translator.tex
\begin{tikzpicture}[auto,>=latex']
    \input{Figs/sim_to_real/tikzset}

    \node[inner sep=0pt] (img_sim) {\includegraphics[width=3cm]{Figs/sim_to_real/sim.png}};
    \node [enc, sim, below=1.5em of img_sim, transform shape] (enc_sim) {$\phi$};
    \node [outer sep=0, inner sep=0, below=1.5em of enc_sim] (loss_eq) {$\lVert z^\mathrm{real} - z^\mathrm{sim} \rVert^2$};
    \node [enc, real, inner sep=0.3em, trapezium angle=75, below=1.5em of loss_eq, transform shape] (enc_real) {$\hat{\phi}$};
    \node[inner sep=0pt, below=1.5em of enc_real] (img_real) {\includegraphics[width=3cm]{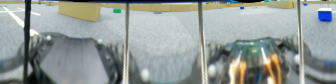}};

    \draw[->](img_sim.south) -- node[node_label, pos=0.5,right] {$o^{\mathrm{sim}}$} (enc_sim.north);
    \draw[->](enc_sim.south) -- node[node_label, pos=0.5,right] {$z^{\mathrm{sim}}$} (loss_eq.north);

    \draw[->](img_real.north) -- node[node_label, pos=0.5,right] {$o^{\mathrm{real}}$} (enc_real.south);
    \draw[->](enc_real.north) -- node[node_label, pos=0.5,right] {$z^{\mathrm{real}}$} (loss_eq.south);

\end{tikzpicture}

%% file: Figs/sim_to_real/real_world_deployment.tex
\begin{tikzpicture}[auto,>=latex']
    \input{Figs/sim_to_real/tikzset}
    \node[inner sep=0pt] (enc_i_img) {\includegraphics[width=3cm]{Figs/sim_to_real/real.png}};
    \node [enc, real, below=1.5em of enc_i_img, transform shape] (enc_i_enc) {$\hat{\phi}$};
    \node [below left=1.0em and 2em of enc_i_enc.south] (node_j_msg) {$z_t^j$};
    \node [block, sim, below=2.25em of enc_i_enc] (enc_i_gnn) {$\theta$};
    \node [block, sim, below=2.25em of enc_i_gnn] (enc_i_psi) {$\psi$};
    \node [coordinate, below=1.0em of enc_i_psi] (enc_i_action) {};

    \draw[->](enc_i_img.south) -- node[node_label, pos=0.5,right] {$o_t^i$} (enc_i_enc.north);
    \draw[->](enc_i_enc.south) -- node[node_label, pos=0.5,right] {$z_t^i$} (enc_i_gnn.north);
    \draw[comm, ->](node_j_msg) -- (enc_i_gnn.west);
    \draw[->](enc_i_gnn.south) -- node[node_label, pos=0.5,right] {$x_t^i$} (enc_i_psi.north);
    \draw[->](enc_i_psi.south) -- node[node_label, pos=1.0,right] {$u_t^i$} (enc_i_action);

\end{tikzpicture}